\documentclass[letterpaper, 10 pt, conference]{ieeeconf}  

\IEEEoverridecommandlockouts                              

\usepackage{graphicx}
\usepackage{subcaption}
\usepackage{multicol}
\usepackage{url}
\usepackage{hyperref}

\title{\LARGE \bf
Commodifying Pointing in HRI:\\Simple and Fast Pointing Gesture Detection from RGB-D Images 
}

\author{Bita Azari$$, Angelica Lim$$ and Richard T. Vaughan$^{*}$
\thanks{*Autonomy Lab, School of Computing Science, Simon Fraser University
{\tt\small\{bazari, angelica, vaughan\}@sfu.ca}}}

\begin{document}

\maketitle
\thispagestyle{empty}
\pagestyle{empty}

\begin{abstract}

We present and characterize a simple method for detecting pointing gestures suitable for human-robot interaction applications using a commodity RGB-D camera. We exploit a state-of-the-art Deep CNN-based detector to find hands and faces in RGB images, then examine the corresponding depth channel pixels to obtain full 3D pointing vectors. We test several methods of estimating the hand end-point of the pointing vector. The system runs at better than 30Hz on commodity hardware: exceeding the frame rate of typical RGB-D sensors. An estimate of the absolute pointing accuracy is found empirically by comparison with ground-truth data from a VICON motion-capture system, and the useful interaction volume established. Finally  we show an end-to-end test where a robot estimates where the pointing vector intersects the ground plane, and report the accuracy obtained. We provide source code as a ROS node, with the intention of contributing a commodity implementation of this common component in HRI systems.
\end{abstract}
\section{INTRODUCTION}
Human-Robot Interaction researchers and developers often seek interaction methods that are quick, intuitive and require little user training. Hand gestures are popular, due to their familiarity from everyday human-human interaction. Of these, pointing is a canonical gesture, used to direct the attention of an interaction partner to an object or place of interest. Pointing is easy for users to do, and easy for human observers to understand. Many authors have used pointing gestures in HRI systems, so we consider the ability to quickly, reliably and accurately detect pointing gestures as an important tool in the HRI repertoire. The goal of this work is to provide a practical, reusable implementation of this component to the community, and to describe its performance. 

We describe a simple and robust pointing gesture recognition system that detects pointing gestures in individual RGB-D frames at $>30$ frames per second on commodity hardware. We exploit a state-of-the-art Deep CNN-based detector to find hands and faces in RGB images, then examine the corresponding depth channel pixels to obtain full 3D pointing vectors. We test several methods of estimating the hand end-point of the pointing vector. An estimate of the absolute pointing accuracy is found empirically by comparison with ground-truth data from a VICON motion-capture system, and the useful interaction volume established. Finally  we show an end-to-end test where a robot estimates where the pointing vector intersects the ground plane, and report the accuracy obtained. We provide source code as a ROS node, with the intention of contributing a commodity Open Source implementation of this common component of HRI systems, with state of the art robustness.


Most gesture-based HRI systems explore only close-range face-to-face interactions. We aim to provide as large a usable interaction volume as we can given the sensor capabilities. We consider the dominant version of pointing, where the vector to be communicated originates at the pointer's eye and passes through the end of the pointing  finger.  The challenge here is that the human hand is a small, deformable object and hard to detect against cluttered backgrounds at long distances. Faces are easier to detect as they are natural fiducials, larger and less deformable than hands. 

The best currently available methods for hand and face detection are based on machine learning, where feature extraction and object detection proposals are learned from labeled data in an end-to-end process. In particular, deep Convolutional Neural Networks (CNNs) have demonstrated tremendous success in object detection tasks.  We employ a custom state of the art CNN model to detect image regions corresponding to hands and faces in 2D RGB images reliably at up to 10m from the camera, where hands have just a few pixels. In this paper we consider how to use the 2D hand and face regions and corresponding depth pixels to robustly estimate the intended 3D pointing vector. We obtain a usable range of around 5m from a camera with 60 degree field of view. This usable interaction volume is large enough for both table-top and mobile robot interactions. 

\begin{figure}[t]
  \centering
  \begin{subfigure}{.49\columnwidth}
    \centering
    \includegraphics[width=\linewidth]{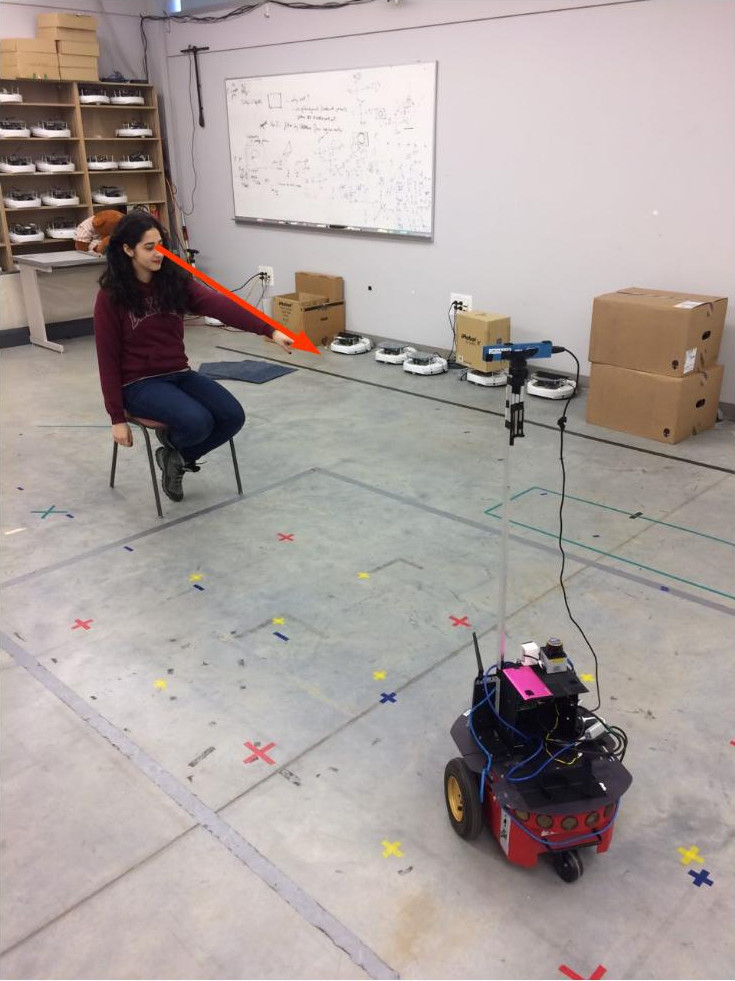}
  \end{subfigure}%
  \hfill
  \begin{subfigure}{.49\columnwidth}
    \centering
    \includegraphics[width=\linewidth]{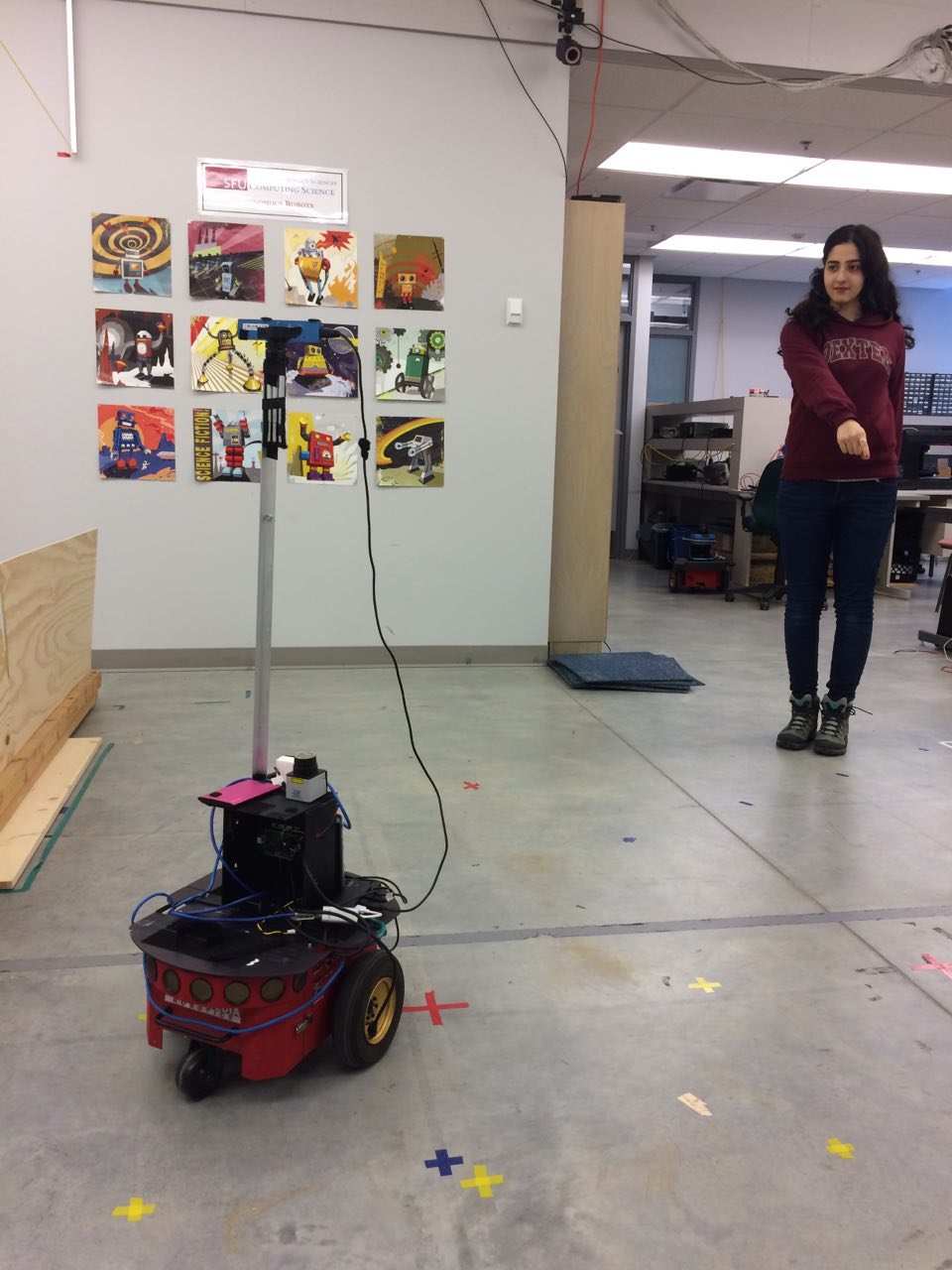}
  \end{subfigure}%
  \caption{Pointing gesture detection for human-robot interaction}
  \label{intro_fig}
\end{figure}

\section{Related Work}

There have been many studies using pointing gesture detection for human-robot interaction. In terms of capturing human body and postures, various approaches were proposed. Pointing gesture detection began with the help of wearable devices, like glove-based devices \cite{quam1990gesture,dipietro2008survey}. With recent achievements in computer vision, a new era began; gesture recognition using vision-based methods are reviewed in \cite{mitra2007gesture} and particularly hand gesture recognition \cite{suarez2012hand,rautaray2015vision}. In vision-based methods, the camera plays an important role: stereo camera, multi cameras, Time-Of-Flight(TOF) camera or depth camera are different approaches for solving pointing gesture detection. \cite{watanabe2000detection,kehl2004real} proposed multi-camera approaches which are promising but less convenient for mobile-robot HRI. A TOF camera was used in \cite{droeschel2011learning}.

\begin{figure}[ht]
  \centering
  \begin{subfigure}{.5\columnwidth}
    \centering
    \includegraphics[width=\linewidth]{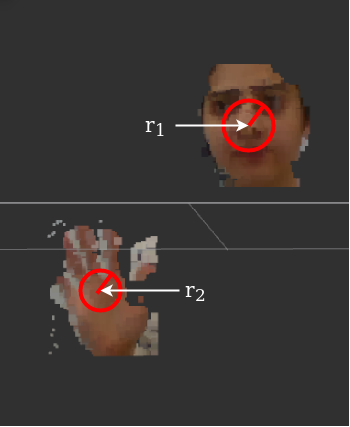}
    \caption{}
    \label{CoBB}
  \end{subfigure}%
  \hfill
  \begin{subfigure}{1\columnwidth}
    \centering
    \includegraphics[width=\linewidth]{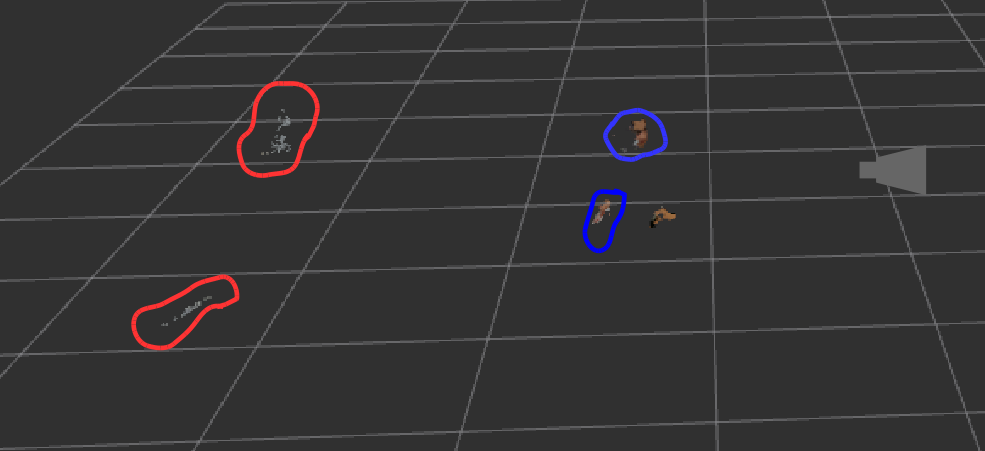}
    \caption{}
    \label{cluster}
  \end{subfigure}%
  \hfill
    \caption{(a) CoBB method applied on 2D image. (b) DBSCAN clusters' visualization in RVIZ; Blue clusters will be selected and red ones are background data which will be eliminated }
\end{figure}

Various methods have been used to detect pointing gestures in these sensor data,  including parametric Hidden Markov Models (HMMs) \cite{wilson1999parametric}. Cascade HMMs \cite{park2011real} give good results in stereo camera data, but obtaining high accuracy in this method depends on a large number of HMM states in the first stage which requires relatively long processing time and lots of training data. In \cite{nickel2003pointing,nickel2004real,nickel2007visual} a neural network was used to detect head pose, and an HMM-based approach was used for pointing gesture recognition; however, their approach uses a three feature sequence to detect pointing gesture that causes delayed detection. Our approach is simpler, and detects pointing in single frames. More recently  \cite{richarz2007monocular} used a neural network architecture which is sensitive to pose variations and the pose that it was trained with. In \cite{pateraki2014visual}, prior knowledge of a limited set of pointing goal points were assumed, and pointing gesture direction was only detected in the horizontal axis. \cite{cosgun2015did,jing2013human,ueno2014efficient,wittner2015s, lai20163d} used the skeleton tracker  provided by the Kinect NITE library which struggles with occluded body parts, it also performs poorly at very close distances. \cite{shukla2015probabilistic} proposed a probabilistic appearance-based model trained with images captured from different viewpoints which is independent of the user's body posture and do not require full-body or partial-body postures for detection, however it relies on hand and finger pose which are only available at close ranges since lots of hand-pixels are required. 

\begin{figure}[t]
\centering
{\includegraphics[width=0.5\textwidth]{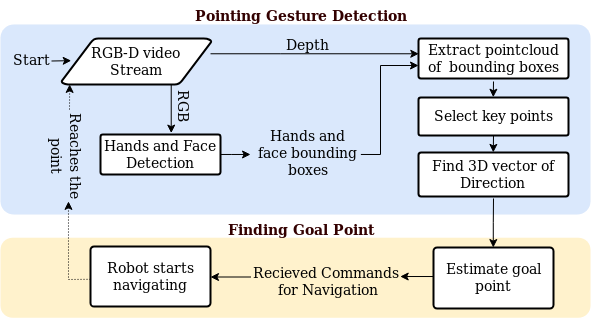}}
\caption{System Flowchart}
\label{system_flowchart}
\end{figure}

\section{Method}
We use the common Intel Realsense ZR300 stereoscopic depth-sensing camera. The manufacturer claims the usable depth capture range of the sensor is from 0.55m to 2.8m. We were able to obtain reasonably accurate pointing vectors at longer ranges than this in practice, up to 5m. In the following, we describe how we address this challenge and we describe each component of our pointing gesture detection system in detail.\\
Our proposed system for pointing gesture detection can be described in two stages: { \it A)} hand and face detection, { \it B)} pointing gesture detection.

\subsection{Hand and Face Detection} 

\begin{figure}[t]
  \centering
  \begin{subfigure}{.5\columnwidth}
    \centering
    \includegraphics[width=\linewidth]{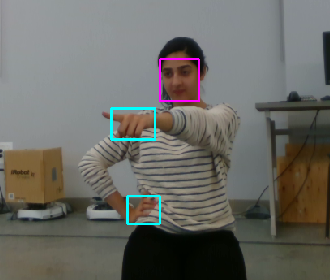}
    \caption{Bounding boxes}
    \label{bounding_box}
  \end{subfigure}%
  \hfill
  \begin{subfigure}{.5\columnwidth}
    \centering
    \includegraphics[width=\linewidth]{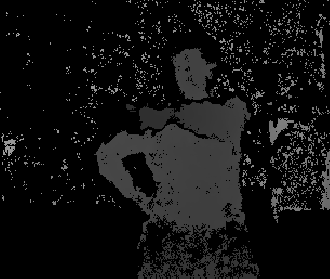}
    \caption{Corresponding depth image}
    \label{depth_image}
  \end{subfigure}%
  \hfill
  \begin{subfigure}{.5\columnwidth}
    \centering
    \includegraphics[width=\linewidth]{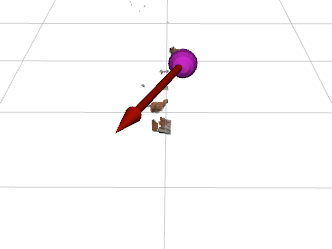}
    \caption{3D vector of pointing gesture}
    \label{3d_vector}
  \end{subfigure}
  \caption{Pointing gesture detection steps}
\end{figure}

In our computations for detecting pointing gestures, we require one pointing hand and the face of a user, which allow us detect pointing gestures even when other parts of the body are occluded. 
We employ a robust state-of-the-art hands-and-face detector based on a deep Convolutional Neural Network (CNN) derived from YOLOv2 \cite{redmon2016yolo9000}. YOLOv2 is an object recognition system which generates object detection candidate bounding boxes, with a probability of the presence of each object label. This network was modified and re-trained by hand-labeled data for face and hand detection that is so reliable and robust that we  obtain stable hand and face detections in almost all frames. Detection is done in every frame, with about 15msec computation time on a commodity GPU (NVIDIA GTX 970). False positives are very rare. For increased paranoid-level robustness and smoothing we apply a bank of Kalman filters to track detections from frame to frame at 30Hz. The output of this system is Region of Interest (ROI) corresponding to each hand and detected face in 2D image. Fig. \ref{bounding_box} shows an example of detected hands and face provided by the network. When a person is in field of view of camera, their face and hands are very reliably located in the RGB image. The CNN model and its training is described in detail in another paper (MohaimanianPour \& Vaughan, submitted to IROS 2018). 

\begin{figure*}[t]
  \centering
  \subcaptionbox{Closest Point to camera}[.3\linewidth][c]{%
    \includegraphics[width=.3\linewidth]{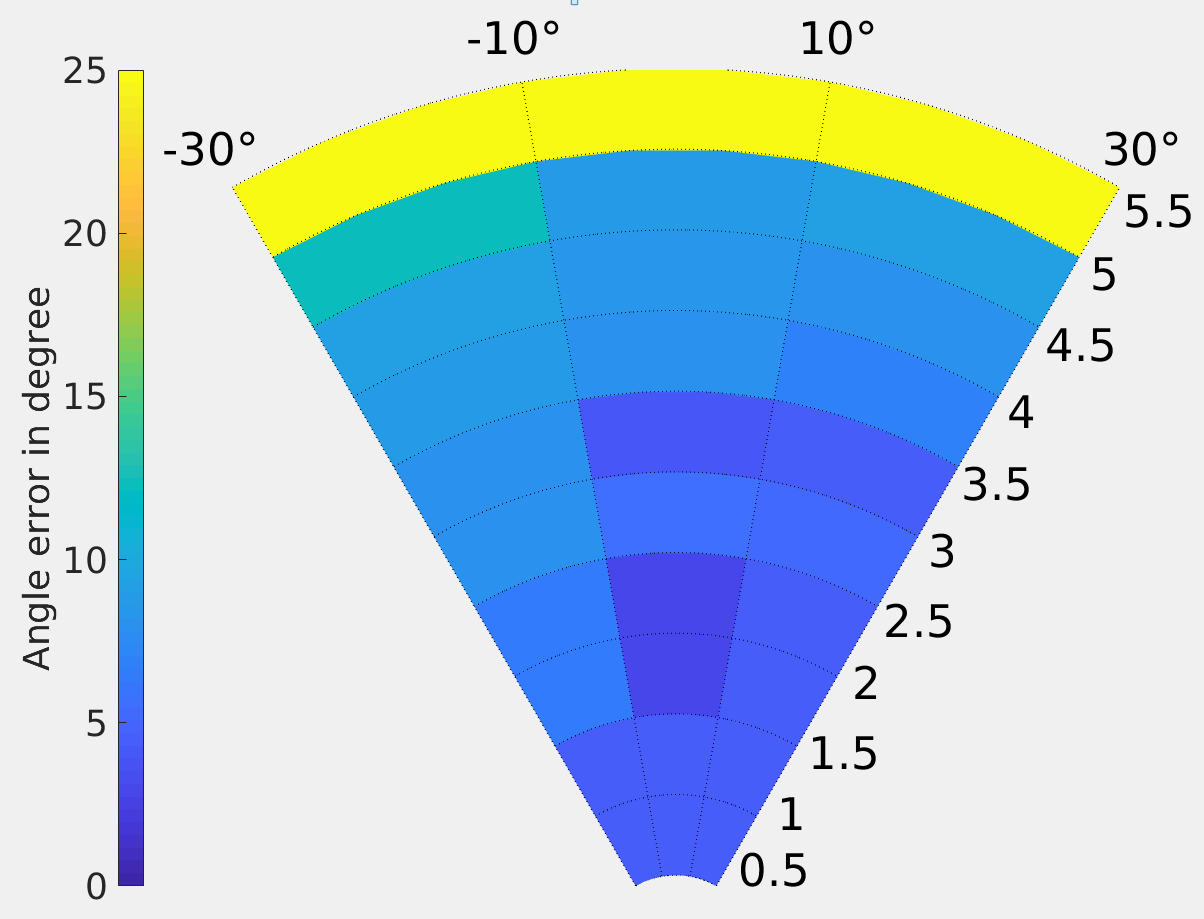}
    }
  \subcaptionbox{Mean of depth}[.3\linewidth][c]{%
    \includegraphics[width=.3\linewidth]{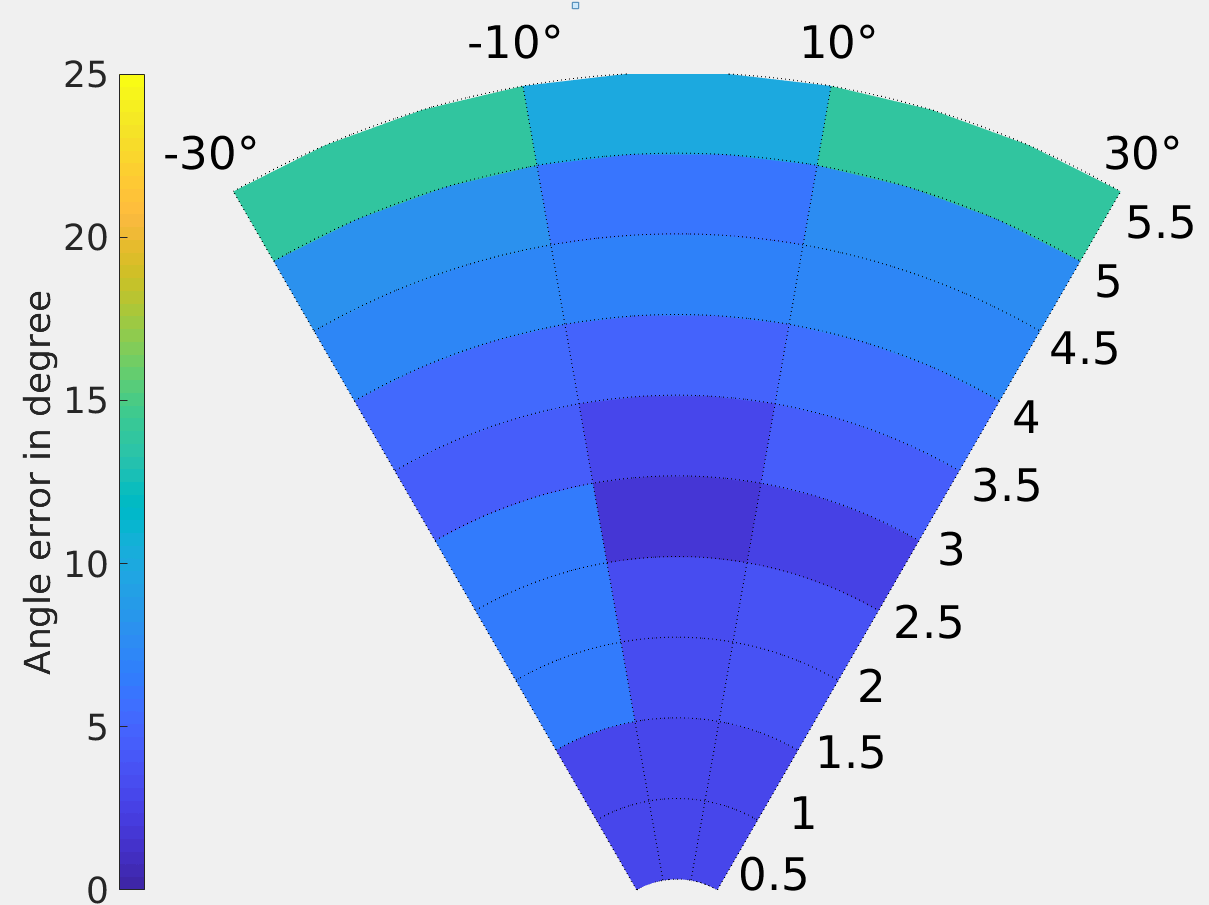}
    }
  \subcaptionbox{Median of depth}[.3\linewidth][c]{%
    \includegraphics[width=.3\linewidth]{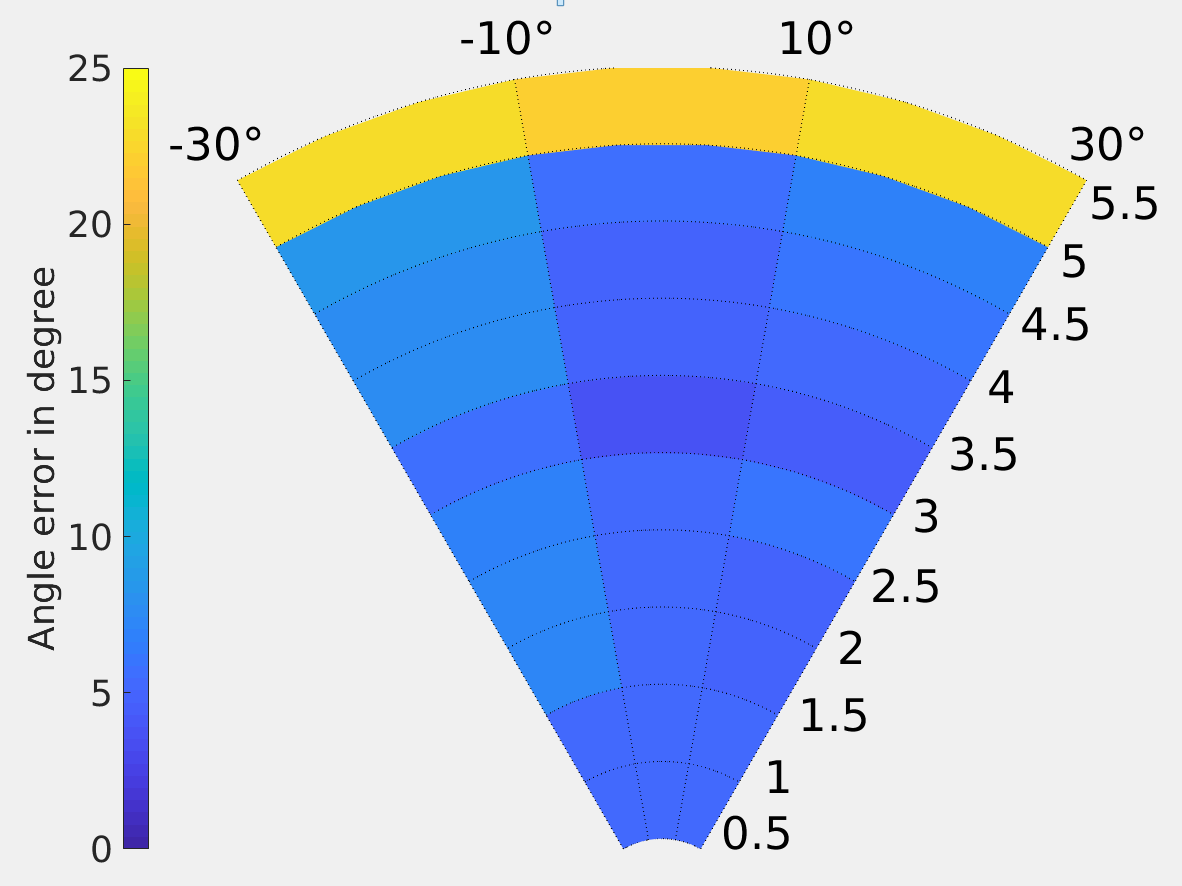}
    }
    \caption{Pointing accuracy obtained with alternative keypoint selection strategies (radius is distance from camera in meters) }
     \label{results}
\end{figure*}    

\begin{figure}[t]
\centering
{\includegraphics[width=0.5\textwidth]{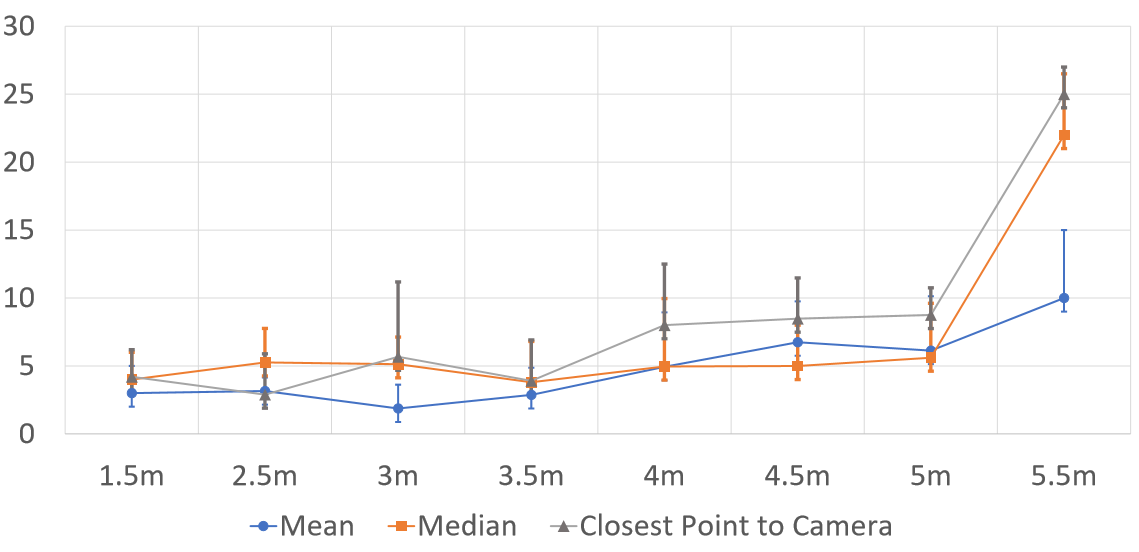}}
\caption{Angular Error Analysis }
\label{diagram}
\end{figure}

\subsection{Pointing Gesture Detection}
After detecting hands and face in 2D, we consider corresponding locations of the ROI in the depth image (since the RGB and depth images are pixel-aligned). When two hands are present in a frame, we must choose one and assume it is performing the pointing gesture. As a simple heuristic, we use the hand closest to the top of the image as is natural in most cases. More interesting heuristics are easily substituted. 

Given pixel locations and depth, and calibrated camera intrinsics, we obtain point clouds for each region that hopefully contain points on either the hand or face. Since hands are small, we typically have only a few sparse points for the pointing-hand. 

Then, we  estimate the intended pointing vector in 3D starting from some point in vicinity of the head and passing through some point in the vicinity of the hand. These two keypoints must be selected, but it is not obvious how to decide them.  This method is similar to the 2D method using the eye-finger line described in \cite{couture2010selecting}, but here in 3D.  
The hand and face bounding boxes often contain depth points that are either from the background behind the user, and spurious depth estimates from the imperfect sensor. We need to eliminate these irrelevant points. We address this problem with two simple but effective complementary approaches: 1) considering only points that lie close to the center of the bounding box in 2D image (CoBB); and 2) clustering points in depth using DBSCAN (Density-based spatial clustering of applications with noise) \cite{ester1996density} to eliminate outliers. In the following section, we describe the methods in detail.
\subsubsection{Considering the Center of the Bounding Boxes(CoBB)}
The first approach is a simple geometrical heuristic. It is naive but fast. The output of the CNN model usually has hands and face located in the center of their detected bounding boxes, and that the bounding boxes are slightly larger than the objects they contain. Thus pixels far from the center of the bounding boxes are more likely to be background or spurious points, and points close to the center are more likely to be correct hand or face target hits. To filter out the unwanted points  we consider a circle with radius \textbf{r} in the center of bounding boxes, and ignore the points outside of the circle as shown in Fig. \ref{CoBB}. Parameter \textbf{r} was hand-tuned by experiment: it has to be large enough so that it could contain some points at large distances where depth data is very sparse, but small enough to eliminate background pixels with high probability. We found that \textbf{r} $= 35 \%$ of the smaller of the width and height of the rectangle worked well. This simple and fast (O(n)) technique serves to filter out most background and spurious points. 

Having removed most outliers, we find the two keypoints that define the pointing vector based on the remaining pointclouds. The angle to the keypoint is obtained  by the geometric mean in pixel-space $(p,q)$ and depth of the clustered points, projected using the camera intrinsics. To find the depth We compared three straightforward approaches: 1) mean, 2) median and 3) minimum depth to the camera in each of the face and hand regions. 

\subsubsection{DBSCAN}
The second approach uses explicit clustering in an effort to improve performance with relatively poor data at large distances from the camera. As explained earlier, the camera claims good quality depth perception up to around 2.8m. At greater distances the error increases rapidly, and very few accurate points are returned. DBSCAN is a popular robust generic clustering technique. We apply it to find clusters of points that correspond to the target, and reject spurious and background points. DBSCAN clustering can find arbitrary numbers of arbitrary shaped clusters, each with a specified minimum number of points, and reject outliers. In this algorithm, a cluster is defined by the density and connectivity of data points; any point that is not part of a cluster is rejected as noise. We used DBSCAN to cluster points according to their depth, to reliably obtain a foreground/target cluster, plus a background cluster, and reject outlier spurious depths (Fig.~\ref{cluster}). We assume the cluster with highest number of points to be the hands and face and other clusters as background and noise. Alternatively we could take the closest cluster to the camera.

Finally, the keypoint is computed as the geometric mean in pixel-space $(p,q)$ and depth of the clustered points.

\begin{figure}[t]
  \centering
  \begin{subfigure}{.48\columnwidth}
    \centering
    \includegraphics[width=\linewidth]{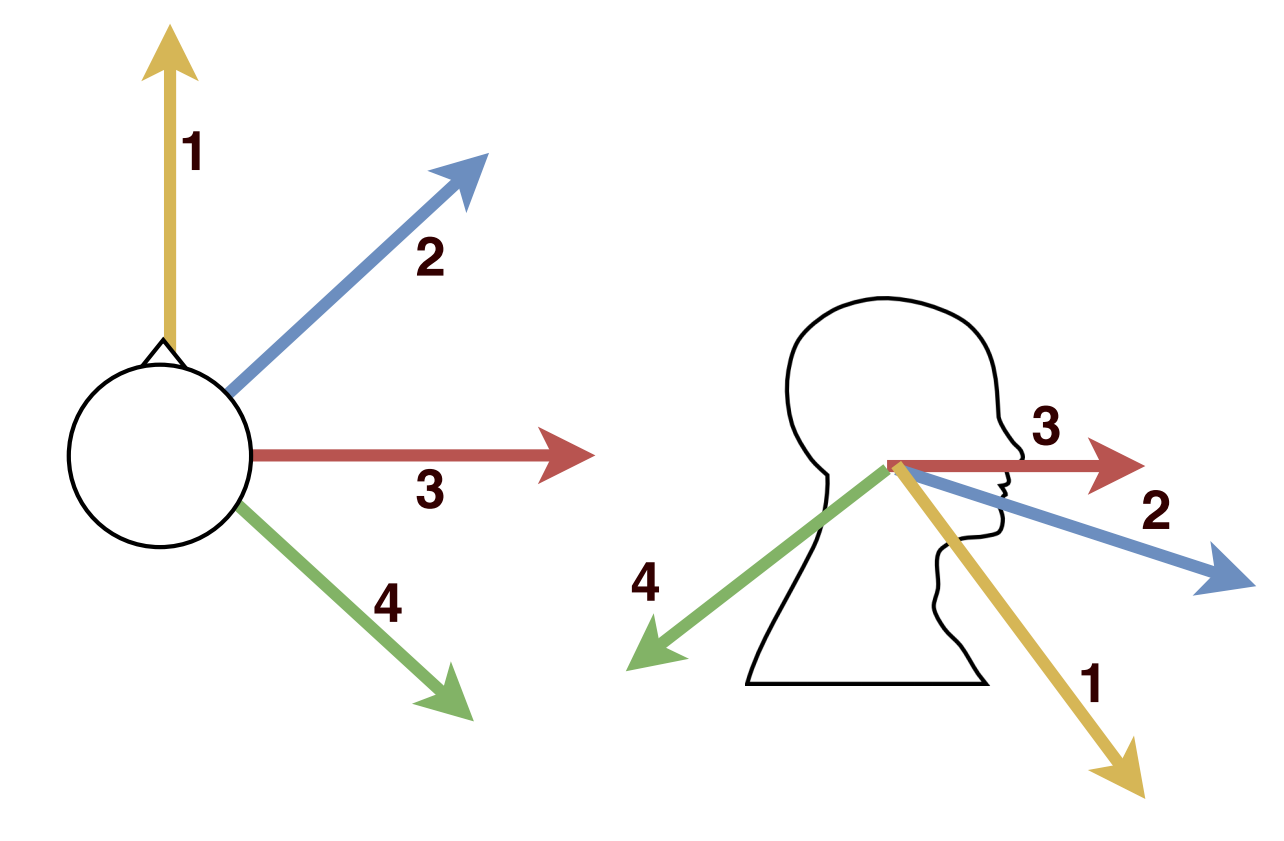}
    \caption{The 4 pointing directions performed in Experiment A. On the left, the approximate pitch of pointing gesture is shown. On the right, we depict the approximate yaw.}
    \label{pointing_dir}
  \end{subfigure}%
  \hfill
  \begin{subfigure}{.48\columnwidth}
    \centering
    \includegraphics[width=\linewidth]{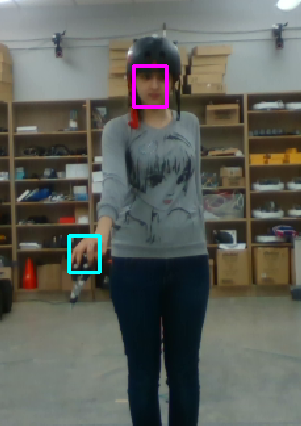}
    \caption{VICON motion capture as ground truth for our experiment\\}
    \label{mocap}
  \end{subfigure}%
  \hfill
  \begin{subfigure}{.8\columnwidth}
    \centering
    \includegraphics[width=\linewidth]{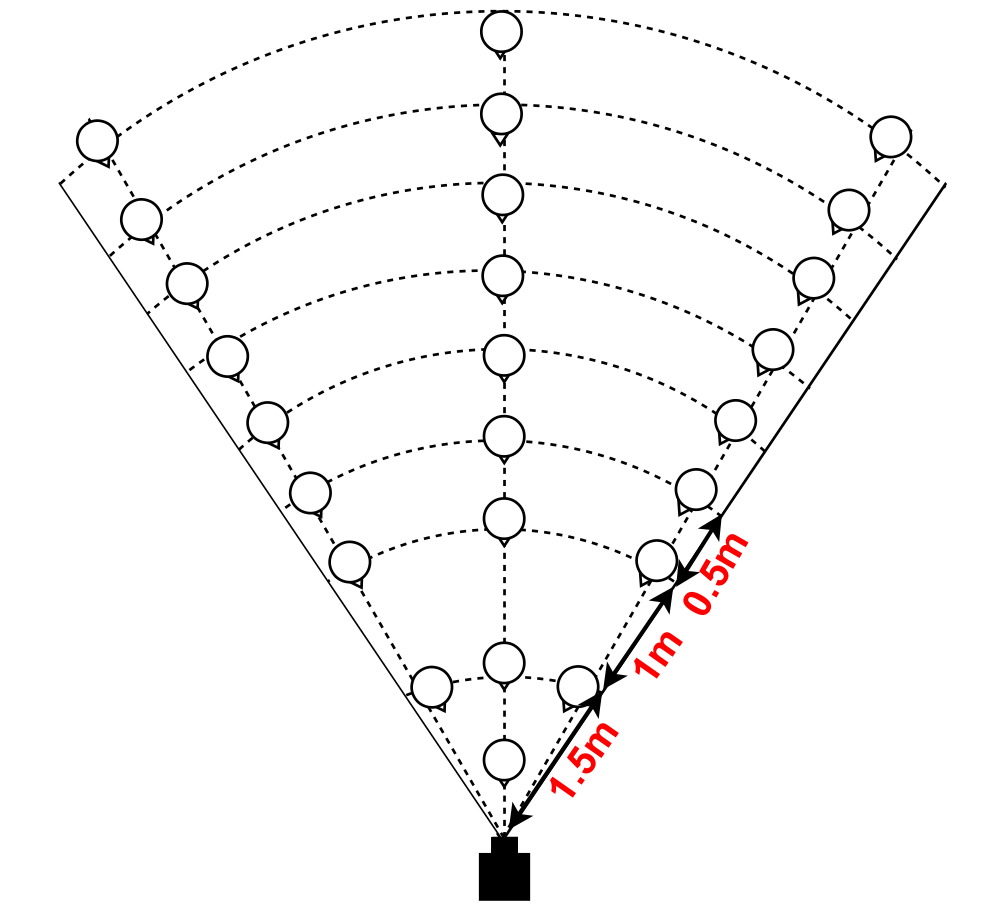}
    \caption{Setting of 25 positions with respect to camera}
  \end{subfigure}
  \caption{Experiment A}
  \label{setting}
\end{figure}

\section{Experiments}

To evaluate our approach, we performed two different experiments. 

First, we measure a lower bound on the pointing vector accuracy by comparing the vector obtained by our vision system with ground truth obtained by an VICON motion capture system. This evaluates how well we estimate the hand and face positions, agnostic to the user's intended pointing target. Two participants with different heights stand in 27 pre-set poses relative to the camera, covering it's practical field of view. At each pose, the user points for two seconds to each of four pre-set directions. The user holds small rigid pointer, hardly visible to the RGB-D camera, and wearing a tight-fitting helmet, both marked with VICON markers. Figure \ref{setting} shows these settings. 

The second experiment is an end-to-end test to validate the suitability of the approach for HRI applications. At each of the trial distances from the camera, the user points to three fixed targets marked on the floor of the lab. The robot finds the intersection of the detected pointing vector with the ground plane to determine where in the lab the user is pointing. We report the ground truth error distance between the point on the floor that the user intended to point to and the robot's estimate of this. This evaluates how well the entire pointing system works: both the eye-to-hand vector approach as well as our specific vision system. The experiment was repeated for two users of different heights.

\begin{figure}[ht]
\centering
{\includegraphics[width=0.4\textwidth]{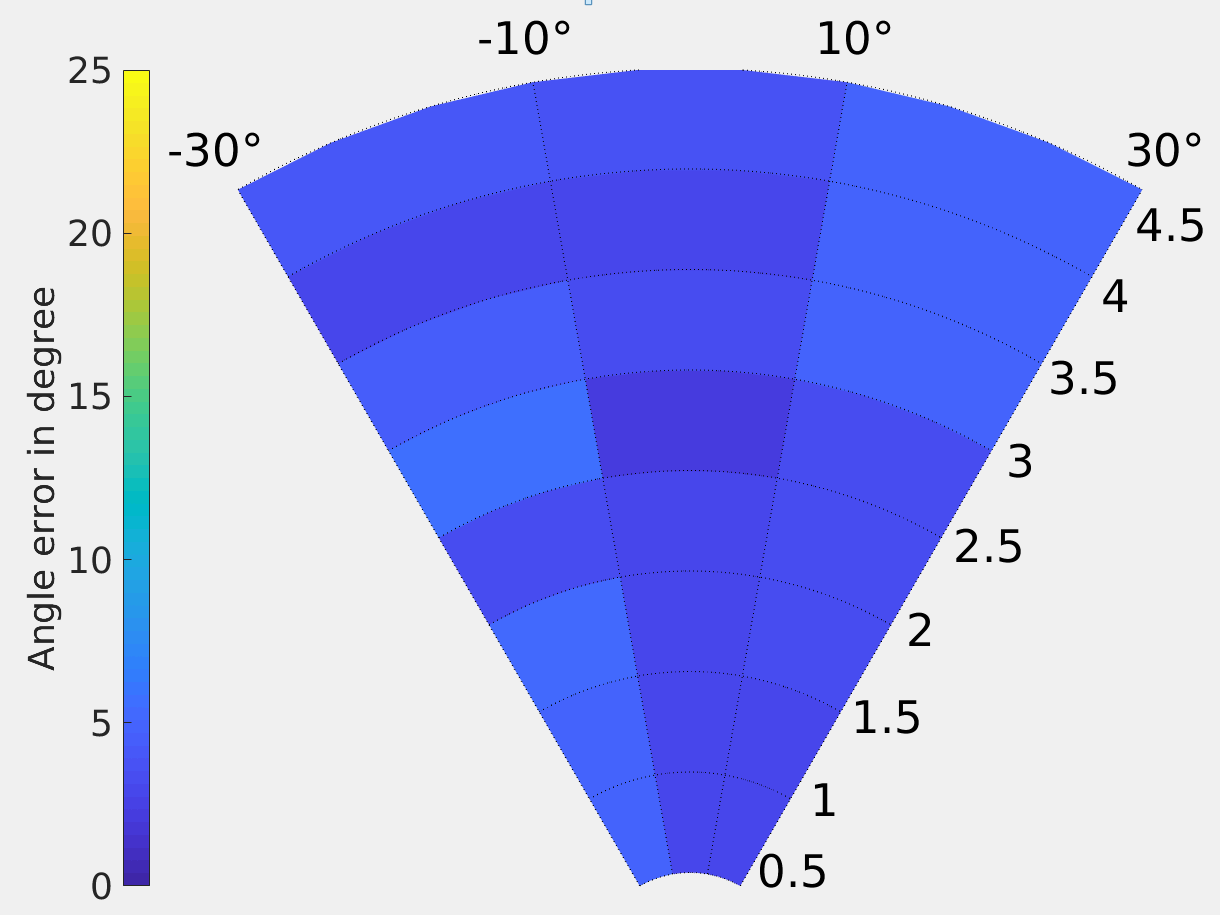}}
\caption{DBSCAN clustering results (radius in meters)}
\label{dbscan}
\end{figure}

\begin{figure*}[t]
\begin{multicols}{3}
    \includegraphics[width=\linewidth]{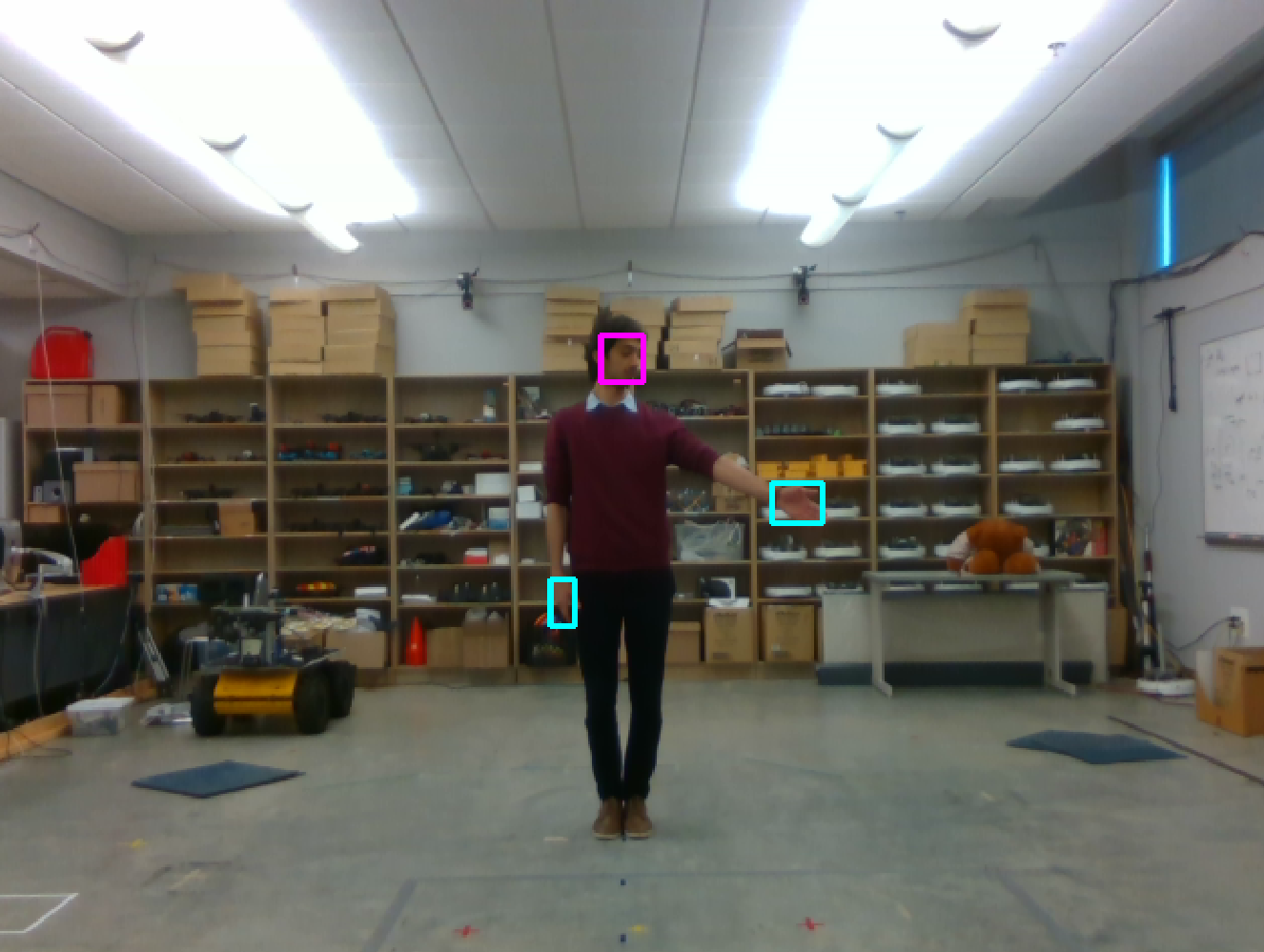}\par 
    \includegraphics[width=\linewidth]{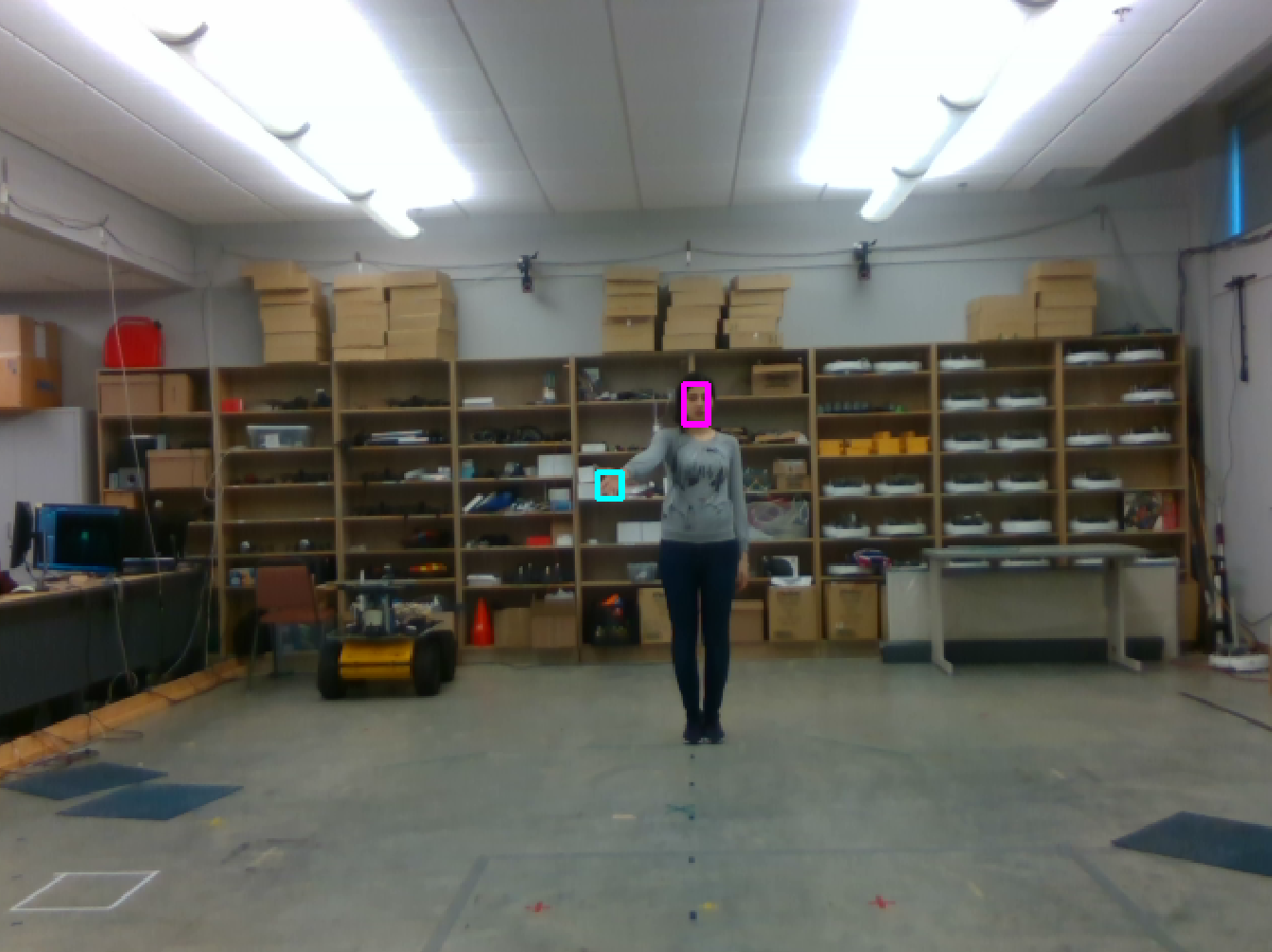}\par
    \includegraphics[width=\linewidth]{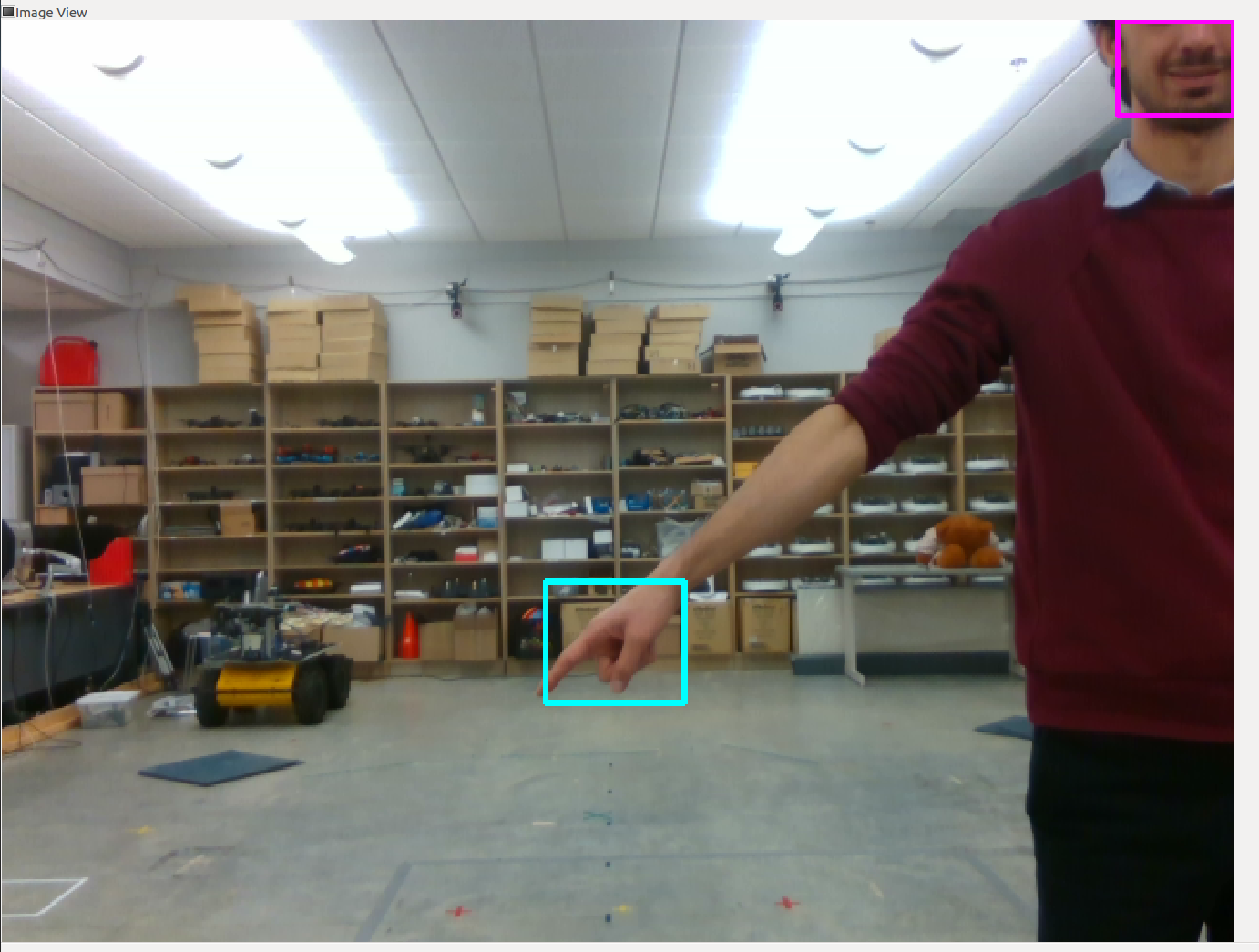}\par
        \end{multicols}
        \caption{Distance and pointing gesture change in our evaluation}
\end{figure*}

\subsection{Experiment A: Pointing Angle Accuracy Compared to VICON}

In our first trial, we evaluated our system using the CoBB method. To retrieve positions of hands and face using a keypoint, we compared the results of using the a) mean, b) median of the points or c) simply picking the point closest to the camera. Two participants were asked to stand in 25 positions and point to 4 different directions, including pointing away from the camera, as shown in Fig. \ref{pointing_dir}. We tested our system up to 5.5 meters from camera and +/- 60 degrees as the horizontal field of view of the Intel Realsense ZR300 camera is 68 degrees. The sample points are shown in Fig. \ref{setting}. 

The results of this experiments were gathered and analyzed by comparing our result (pitch and yaw of hands with respect to the face), with the ground truth (pitch and yaw calculated from VICON markers). The error is reported as the mean of differences in angle. The results are shown in Fig.~\ref{results} and Fig.~\ref{diagram}.  The measured angular error varied from 0 to 10 degrees, increasing as the user is further from the camera. The results show that calculating keypoints using the mean depth of inlier depth points was more accurate, especially further from the camera. 

Results of the DBSCAN clustering approach are reported in Fig.~\ref{dbscan}. Similar to other experiments, it was performed with two persons, in 25 positions and performing 4 pointing gestures. The results show that DBSCAN clustering improves accuracy compared to the geometric CoBB approach. However, at larger distances the clustering method fails to detect a cluster in many video frames, since the sensor does not provide enough points. At distances below 4m we obtained clusters and thus pointing vectors in almost every frame at 30Hz. At distances of 4m to 5m, due to depth data being only occasionally available, we obtain pointing vectors at around 10Hz. Again, the camera specification says depth data is good to only 2.8m.

\subsection{Experiment B: Pointing to targets on floor }
Since one of the applications of pointing gesture detections is for commanding robot to reach to a goal point, we validate our system using marked points on the ground with known position with respect to camera. To estimate goal points the intersection of the 3D vector of pointing gesture with ground is calculated. This is achieved by intersecting the pointing vector with the ground plane at $z=0$:

\begin{equation}\label{robot_height_eq}
\begin{array}{l}
	Z_{face}' = Z_{camera} + Z_{face}, \\
    Z_{hand}' = Z_{camera} + Z_{hand}
\end{array}
\end{equation}

\begin{equation}\label{plane_eq}
	\vec{P} = \vec{(face-hand)}
\end{equation}

\begin{equation}\label{coeff_eq}
	t = \frac{Z_{face}'}{P_z}
\end{equation}

\begin{equation}\label{goal_point_eq}
	Goal   = (X_{face} - (t * P_x), Y_{face} - (t * P_y), 0)
\end{equation}

As ground truth, the target points on the floor were measured by hand and marked with tape. In this experiment, two persons were asked to stand in 5 different positions, pointing to 3 marked points on the ground, goal point detected by our system was compared with known positions using euclidean distances between goal point and marked point. Results are shown in Table \ref{table:goal_point}. The variance of error detection was between 0.008 to 0.01, which shows robustness of the system.

Having detected a target point on the floor, we complete our end-to-end robot system by having the robot navigate to the target. The robot 
measures the covariance in the pointing direction estimate for the most recent 30 frames (1 second). If the covariances collects a batch of 30 frames in 1 second, and if the covariances fall below a threshold, the robot navigates to the indicated goal point. After reaching the goal point it will then start observing around to find other pointing gesture. For the work described in here we use Pioneer DX3 robot as shown in Fig. \ref{intro_fig} with the Realsense ZR300 RGB-D camera connected to a laptop with a commodity GPU (NVIDIA GeForce GTX 1060) which provides the computational power needed for running the deep neural network for hands and face detection. The mobile robot base is controlled by the built-in computer running ROS (Robot Operating System). 

These results are comparable in accuracy at close ranges to other reported systems listed in \cite{shukla2015probabilistic}. However, our system has a larger usable envelope, and runs at the frame rate of the RGB-D sensor. 

\begin{table}[ht]
 \centering
 \caption{}
 \begin{tabular}{lcccc}
 \hline
  Distance (m) & $\mu$ (cm) & $\sigma$ (cm) \\
 \hline
  1.5 & 16.1 & 1.9 \\
  2.5 & 18.1 & 2.1 \\
  3.5 & 14.5 & 3.5 \\
  4.5 & 22.4 & 5.6 \\
  5.5 & 48.4 & 12.3 \\
 \hline
 \end{tabular}
 \label{table:goal_point}
\end{table}

\section{CONCLUSIONS}
The goal of this work is to provide a practical, robust approach to pointing for HRI applications, to demonstrate a system and to describe its performance.  

We presented a method for pointing gesture recognition that is able to detect pointing gestures in complex environment with cluttered visual backgrounds and varied lighting, at ranges up to 5.5m.  We compared two approaches for outlier rejection and determining the key points of the pointing vector. A clustering approach using DBSCAN was more accurate but gave results in only a subset of frames when the user was far from the camera. A geometric heuristic gave results more often but with greater error at large distances.

This system is offered as commodity component for HRI systems, with state of the art speed and robustness, comparable performance to other systems at close range, but a larger usable interaction envelope. 

\section*{Code and Reproducibility}
Source code including a ROS node is provided at \url{https://github.com/AutonomyLab/pointing_gesture}. The commit hash for the version used to obtain the results in this paper is 7a4fe3a102c528c606cb3cac6e91cede8d54b80a. 


\bibliographystyle{IEEEtran}
\bibliography{root}

\end{document}